\documentclass[onecolumn,10pt]{article}
\usepackage{spconf}
\usepackage[utf8]{inputenc}

\usepackage{cite}
\usepackage{fixltx2e}

\usepackage{graphicx}

\usepackage{subcaption}              

\usepackage[utf8]{inputenc}

\usepackage{amsfonts}
\usepackage{amsmath}
\usepackage{mathtools}

\usepackage{amssymb}

\usepackage{bm}

\usepackage{color,verbatim}
\usepackage{multirow}
\usepackage{accents}

\usepackage{theoremref}

\usepackage{flushend}

\usepackage{url}

\newcounter{rulecounter}
\newcommand{\resetrule}{ \setcounter{rulecounter}{0}}
\resetrule

\newsavebox{\selvestebox}
\newenvironment{colbox}[1]
  {\newcommand\colboxcolor{#1}%
   \begin{lrbox}{\selvestebox}%
   \begin{minipage}{\dimexpr\columnwidth-2\fboxsep\relax}}
  {\end{minipage}\end{lrbox}%
   \begin{center}
   \colorbox{\colboxcolor}{\usebox{\selvestebox}}
   \end{center}}

\definecolor{orange}{rgb}{1,0.8,0}
\definecolor{gray}{rgb}{.9,0.9,0.9}
\definecolor{darkgray}{rgb}{.3,0.3,0.3}
\definecolor{darkblue}{rgb}{.1,0.0,0.3}
\definecolor{lightblue}{rgb}{0.7,0.7,1}
\definecolor{lightred}{rgb}{1,0.7,.7}
\definecolor{purple}{RGB}{204,153,255}
\definecolor{lightgray}{rgb}{.95,0.95,0.95}
\definecolor{lightgreen}{rgb}{0.3,0.5,0.3}
\definecolor{darkgreen}{rgb}{0.05,0.3,0.05}

\newcommand{\rfield}{\mathbb{R}}

 \newcommand{\define}{:=}

\newtheorem{myproposition}{Proposition}
\newtheorem{myremark}{Remark}
\newtheorem{myproblemstatement}{Problem Statement}
\newtheorem{mylemma}{Lemma}
\newtheorem{mytheorem}{Theorem}
\newtheorem{mydefinition}{Definition}
\newtheorem{mycorollary}{Corollary}

\usepackage{pgfplots}
\pgfplotsset{compat=newest}
\usetikzlibrary{plotmarks}
\usetikzlibrary{arrows.meta}
\usepgfplotslibrary{patchplots}
\usepackage{grffile}

\pgfplotsset{plot coordinates/math parser=false}
\newlength\mywidth
\newlength\myheight
\definecolor{mycolor1}{rgb}{0.00000,0.44700,0.74100}%
\definecolor{mycolor2}{rgb}{0.85000,0.32500,0.9800}%
\definecolor{mycolor3}{rgb}{0.92900,0.69400,0.12500}%
\definecolor{mycolor4}{rgb}{0.89400,0.18400,0.15600}%
\definecolor{mycolor5}{rgb}{0.46600,0.67400,0.18800}%
\definecolor{mycolor6}{rgb}{0.30100,0.74500,0.93300}%
\definecolor{mycolor7}{rgb}{0.63500,0.07800,0.18400}%

\pgfplotscreateplotcyclelist{colorlist}{%
color=mycolor1,line width=2.0pt,
solid, every mark/.append style={solid, fill=mycolor1}, mark=*\\%
color=mycolor2,line width=2.0pt,dotted, every mark/.append style={solid, fill=mycolor2}, mark=square*\\%
color=mycolor3,line width=2.0pt,densely dotted, every mark/.append style={solid, fill=mycolor3}, mark=otimes*\\%
color=mycolor4,line width=2.0pt,loosely dotted, every mark/.append style={solid, fill=mycolor4}, mark=triangle*\\%
color=mycolor5,line width=2.0pt,
dashed, every mark/.append style={solid, fill=mycolor5},mark=diamond*\\%
color=mycolor6,line width=2.0pt,loosely dashed, every mark/.append style={solid, fill=mycolor6},mark=*\\%
color=mycolor7,densely dashed, every mark/.append style={solid, fill=mycolor7},line width=2.0pt,mark=square*\\%
dashdotted, every mark/.append style={solid, fill=mycolor7},mark=otimes*\\%
dashdotdotted, every mark/.append style={solid},mark=star\\%
densely dashdotted,every mark/.append style={solid, fill=gray},mark=diamond*\\%
}
\pgfplotscreateplotcyclelist{my black white}{%
solid, every mark/.append style={solid, fill=gray}, mark=*\\%
dotted, every mark/.append style={solid, fill=gray}, mark=square*\\%
densely dotted, every mark/.append style={solid, fill=gray}, mark=otimes*\\%
loosely dotted, every mark/.append style={solid, fill=gray}, mark=triangle*\\%
dashed, every mark/.append style={solid, fill=gray},mark=diamond*\\%
loosely dashed, every mark/.append style={solid, fill=gray},mark=*\\%
densely dashed, every mark/.append style={solid, fill=gray},mark=square*\\%
dashdotted, every mark/.append style={solid, fill=gray},mark=otimes*\\%
dashdotdotted, every mark/.append style={solid},mark=star\\%
densely dashdotted,every mark/.append style={solid, fill=gray},mark=diamond*\\%
}
\newcommand{\legendfontsize}{\normalsize}
\newcommand{\ticklabelfontsize}{\normalsize}
\newcommand{\featvec}{\mathbf{x}}
\newcommand{\dataentry}{X}
\newcommand{\datamatrix}{\mathbf{\dataentry}}
\newcommand{\datatensor}{\underline{\datamatrix}}

\newcommand{\nbrinpfeatures}{F}
\newcommand{\nonlinearity}[1]{\sigma(#1)}
\newcommand{\nbrlayfeatures}{P}
\newcommand{\nbrnodes}{N}
\newcommand{\shiftentry}{S}
\newcommand{\shiftmat}{\mathbf{\shiftentry}}
\newcommand{\shiftind}{i}
\newcommand{\nbrshifts}{I}
\newcommand{\outlayfun}{g}

\newcommand{\nbrlayers}{L}
\newcommand{\transpose}{^\top}

\newcommand{\shiftindp}{\shiftind'}

\newcommand{\nodeindp}{\nodeind'}
\newcommand{\regparsmooth}{\mu_2}
\newcommand{\regparsparse}{\lambda}
\newcommand{\regpargraphsmooth}{\mu_1}

\newcommand{\shiftweightfeatnot}[3]{_{#1#2#3}}
\newcommand{\shiftshiftfeatnot}[3]{_{#1#2#3}}
\newcommand{\nodenodeshiftnot}[3]{_{#1#2#3}}
\newcommand{ \outputlayvec}{\check{\mathbf{z}}}
\newcommand{\nodeshiftfeatnot}[3]{_{#1#2#3}}
\newcommand{\nodeshiftnot}[2]{_{#1#2}}

\newcommand{\featind}{p}
\newcommand{\layernot}[1]{^{(#1)}}
\newcommand{\layerind}{l}
\newcommand{\layfunc}{f}
\newcommand{\weightcol}{\mathbf{w}}
\newcommand{\weightmat}{\mathbf{W}}
\newcommand{\weightshiftentry}{R}
\newcommand{\nodeind}{n}
\newcommand{\mappingfun}{\mathcal{F}}

\newcommand{\weighttensor}{\underline{\weightmat}}
\newcommand{\shifttensor}{\underline{\shiftmat}}

\newcommand{\linoutputlaytensorentry}{\underline{Z}}
\newcommand{\linoutputlaymat}{\mathbf{Z}}
\newcommand{\outputlaymat}{\check{\mathbf{Z}}}
\newcommand{\linoutputlaytensor}{\underline{\linoutputlaymat}}
\newcommand{\outputlaytensor}{\underline{\outputlaymat}}
\newcommand{\outputlaytensorentry}{\underline{\check{Z}}}

\newcommand{\weightshifttensor}{\underline{\mathbf{\weightshiftentry}}}
\newcommand{\paramhidvec}{\bm{\theta}_{z}}
\newcommand{\paraminpvec}{\bm{\theta}_{x}}
\newcommand{\diffusedfeatvec}{\mathbf{h}}
\newcommand{\paramoutvec}{\bm{\theta}_{g}}

\newcommand{\nbrclasses}{K}

\newcommand{\predictionmat}{\hat{\mathbf{Y}}}
\newcommand{\predictionmatentry}{\hat{Y}}
\newcommand{\shiftperturbation}{\underline{\mathbf{O}}_{\shifttensor}}
\newcommand{\featperturbation}{\mathbf{O}_{\datamatrix}}
\newcommand{\neighborhoodset}{\mathcal{N}}

\newcommand{\labelentry}{y}
\newcommand{\labelmat}{\mathbf{\labelmatentry}}
\newcommand{\labelmatentry}{Y}

\newcommand{\regfun}{\rho}

\newcommand{\labeledset}{\mathcal{L}}
\newcommand{\vertexset}{\mathcal{V}}

\usepackage{array}
\usepackage{footnote}
\makesavenoteenv{tabular}
\makesavenoteenv{table}

\title{A Recurrent Graph Neural Network for Multi-relational Data}
\name{Vassilis N. Ioannidis$^{\star }$ 
\qquad Antonio G. Marques$^{\dagger}$ 
\qquad Georgios B. Giannakis$^{\star}$
	\thanks{{ The work in this paper has been supported by USA NSF grants 171141,  1500713, and  1442686, and by the Spanish grants MINECO KLINILYCS (TEC2016-75361-R) and Instituto de Salud Carlos III DTS17/00158.}}}
\address{$^{\star}$ Digital Technology Center and Dept. of ECE, University of Minnesota, Minneapolis, USA \\
$^{\dagger}$ Dept. of Signal Theory and Comms., King Juan Carlos University, Madrid, Spain}
\date{September 2018}

\ninept 
\begin{document}

\maketitle
\begin{abstract}
The era of ``data deluge'' has sparked the interest in graph-based learning 
methods in a number of disciplines such as sociology, biology, neuroscience, or 
engineering. In this paper, we introduce a graph \textit{recurrent} neural 
network (GRNN) for scalable semi-supervised learning from 
\textit{multi-relational data}. Key aspects of the novel GRNN architecture are 
the use of multi-relational graphs, the dynamic adaptation to the different 
relations via learnable weights, and the consideration of graph-based 
regularizers to promote smoothness and alleviate over-parametrization. Our 
ultimate goal is to design a powerful learning architecture able to: discover 
complex and highly non-linear data associations, combine (and select) multiple 
types of relations, and scale gracefully with 
respect to the size of the graph. Numerical tests with real datasets 
corroborate the design goals and illustrate the performance gains relative to 
competing alternatives.
\end{abstract}
\section{Introduction}
A task of major importance in the interplay between machine learning and 
network science is semi-supervised learning (SSL) over graphs. In a nutshell, 
SSL aims at predicting or extrapolating nodal attributes given: i) the values 
of those attributes at a subset of nodes and (possibly) ii) additional 
features at all nodes. A relevant example is protein-to-protein interaction 
networks,  where the proteins (nodes) are associated with specific biological 
functions, thereby facilitating the understanding of pathogenic and 
physiological mechanisms.

While significant progress has been achieved for this problem, most works consider that the relation among the nodal variables is represented by a single graph. This may be inadequate in many contemporary applications, where nodes may engage on multiple types of relations~\cite{kivela2014multilayer}, motivating the generalization of traditional SSL approaches for \emph{single-relational} graphs to \emph{multi-relational} graphs\footnote{{Many works in the literature refer to these graphs as multi-layer graphs.}}. In the particular case of social networks, each layer of the graph could capture a specific form of social interaction, such as friendship, family bonds, or coworker-ties \cite{wasserman1994social}. Albeit their ubiquitous presence, development of SSL methods that account for multi-relational networks is only in its infancy, see, e.g.,~\cite{ioannidis2018multilay,kivela2014multilayer}. 

\vspace{.1cm}
\noindent\textbf{Related work.} A popular approach for graph-based SSL methods 
is to assume that the true labels are ``smooth'' with respect to the underlying 
network structure, which then motivates leveraging the topology of the network 
to propagate the labels and increase classification performance. Graph-induced 
smoothness may be captured by kernels on 
graphs~\cite{belkin2006manifold,ioannidis2018kernellearn};  Gaussian random 
fields \cite{zhu2003semi}; or  low-rank {parametric} models based on the 
eigenvectors of the graph Laplacian or adjacency 
matrices~\cite{shuman2013emerging,marques2015aggregations}. Alternative 
approaches use the graph to embed the nodes in a vector space, and classify the 
points~\cite{weston2012deep,yang2016revisiting,berberidis2018adaptive,seo2016structured}. 
More recently, another line of 
works postulates that the mapping between the input data and the labels is given by a 
neural network (NN) architecture 
that incorporates the structure of the graph \cite{bronstein2017geometric, 
gama2018convolutional, kipf2016semi}. The parameters describing the 
NN are then learned using labeled examples and feature 
vectors, and those parameters are finally used to predict the labels of the 
unobserved nodes. See, e.g., \cite{kipf2016semi}, for state-of-the-art results 
in SSL using a single-relational graph when nodes are accompanied with additional features. 

\vspace{.1cm}
\noindent\textbf{Contributions.} This paper develops a deep learning framework 
for SSL over multi-relational data. The main contributions are i) we postulate 
a (tensor-based) NN architecture that accounts for multi-relational  graphs, 
ii) we define mixing coefficients that capture how the different relations 
affect the desired output and allow those to be learned from the examples, 
which enables identifying the underlying structure of the data; iii) at 
every layer we propose a recurrent feed of the data that broadens the class of 
(graph signal) transformations the NN implements and facilitates the 
diffusion of the features across the graph; and iv) in the training phase 
we consider suitable 
(graph-based) regularizers that avoid overfitting and further capitalize on the 
topology of the data. 


\section{Modeling and problem formulation}
\begin{figure}
    \centering
    \includegraphics[width=\columnwidth]{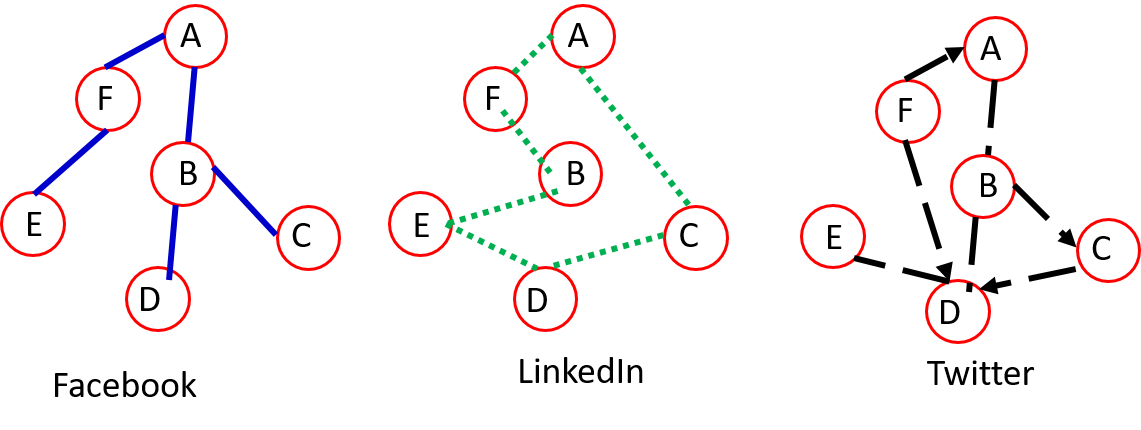}
    \caption{A social multi-relational network of $\nbrnodes=6$ people.}
    \label{fig:multilayer}
\end{figure}

Consider a network of $\nbrnodes$ nodes, with vertex set $\vertexset\define\{v_1,\ldots,v_\nbrnodes\}$, connected through $\nbrshifts$ relations. The connectivity at the $\shiftind$-th relation is captured by  the  $\nbrnodes\times\nbrnodes$ matrix $\shiftmat_\shiftind$, and the scalar $\shiftentry\nodenodeshiftnot{\nodeind}{\nodeindp}{\shiftind}$ represents the influence of $v_\nodeind$ to $v_\nodeindp$ under the $\shiftind$-th relation. In  social networks for example, these may represent the multiple types of connectivity among people  such as Facebook, LinkedIn, and Twitter; see Fig.\ref{fig:multilayer}.  The matrices $\{\shiftmat_\shiftind\}_{\shiftind=1}^\nbrshifts$ are collected in the $\nbrnodes\times\nbrnodes\times\nbrshifts$ tensor $\shifttensor$. The graph-induced neighborhood of $v_\nodeind$ for the $\shiftind$-th relation is 
\begin{align}
\label{eq:neighborhood}
\neighborhoodset_\nodeind^{(\shiftind)}\define\{\nodeindp:\shiftentry\nodenodeshiftnot{\nodeind}{\nodeindp}{\shiftind}\ne0,~~ v_\nodeindp\in\vertexset\}.
\end{align}
We associate an $\nbrinpfeatures\times 1$ feature vector $\featvec_{\nodeind}$ to the $\nodeind$-th node, and collect those vectors in the $\nbrnodes\times\nbrinpfeatures$ feature matrix $\datamatrix\define[\featvec_{1}\transpose,\ldots,\featvec_{\nbrnodes}\transpose]\transpose$. The entry $\dataentry_{\nodeind\featind}$ may denote, for example, the salary of the  $\nodeind$-th individual in the LinkedIn social network.

We also consider that each node $\nodeind$ has  a label of interest $\labelentry_\nodeind\in\{0,\ldots,\nbrclasses-1\}$, which may represent, for example, the education level of a person. In SSL we have access to the labels only at a subset of nodes $\{\labelentry_{\nodeind}\}_{\nodeind\in\labeledset}$, with $\labeledset \subset\vertexset$. This partial availability may be attributed to privacy concerns (medical data); energy considerations  (sensor networks); or unrated items (recommender systems). The $\nbrnodes\times\nbrclasses$ matrix  $\labelmat$ is the ``one-hot'' representation of the true nodal labels, that is, if $\labelentry_\nodeind=k$ then $\labelmatentry_{\nodeind,k}=1$ and $\labelmatentry_{\nodeind,k'}=0, \forall k'\ne k$.

The goal of this paper is to develop a \textit{deep learning architecture} 
based on \textit{multi-relational graphs}  that, using as input the 
features in $\datamatrix$,  maps each node $n$ to a corresponding label 
$\labelentry_\nodeind$ and, hence, estimates the unavailable labels.

\section{Proposed GRNN architecture}
Deep learning architectures typically process the input information using a 
succession of $\nbrlayers$ hidden layers. Each of the layers is composed of a 
conveniently parametrized linear transformation, a scalar nonlinear 
transformation, and, oftentimes, a dimensionality reduction (pooling) operator. 
The intuition is to combine nonlinearly local features to progressively extract 
useful information~\cite{goodfellow2016deep}. GNNs tailor these operations to 
the graph that supports the data \cite{bronstein2017geometric}, including the 
linear \cite{defferrard2016convolutional}, nonlinear 
\cite{defferrard2016convolutional} and pooling 
\cite{gama2018convolutional} operators. In this section, we 
describe the 
architecture of our novel multi-relational GRNN, that inputs the known features 
at the first layer and outputs the predicted labels at the last layer.
We first present the operation of the GNN, GRNN, and output layers, and finally 
discuss the training of our NN.

\subsection{Single layer operation}
Let us consider an intermediate layer (say the $l$th one) of our 
architecture. The output of that layer is the  
$\nbrnodes\times\nbrshifts\times\nbrlayfeatures\layernot{\layerind}$ tensor 
$\outputlaytensor\layernot{\layerind}$ that holds the 
$\nbrlayfeatures\layernot{\layerind}\times 1$  feature vectors 
$\outputlayvec\nodeshiftnot{\nodeind}{\shiftind}\layernot{\layerind}, \forall 
\nodeind,\shiftind$, with $\nbrlayfeatures\layernot{\layerind}$ being the 
number of output features at $\layerind$. Similarly, the  
$\nbrnodes\times\nbrshifts\times\nbrlayfeatures\layernot{\layerind-1}$ tensor 
$\outputlaytensor\layernot{\layerind-1}$ represents the input to the layer.
Since our focus is on predicting labels on all nodes, we do not consider a 
dimensionality reduction (pooling) operator in the intermediate layers. As a 
result, the 
mapping from $\outputlaytensor\layernot{\layerind-1}$ to 
$\outputlaytensor\layernot{\layerind}$ can be split into two steps. First, we 
define a linear transformation that maps the  
$\nbrnodes\times\nbrshifts\times\nbrlayfeatures\layernot{\layerind}$ tensor 
$\outputlaytensor\layernot{\layerind-1}$ into the 
$\nbrnodes\times\nbrshifts\times\nbrlayfeatures\layernot{\layerind}$ tensor 
$\linoutputlaytensor\layernot{\layerind}$.  The intermediate feature output 
$\linoutputlaytensor\layernot{\layerind}$ is then processed elementwise using a 
scalar 
nonlinear transformation $\nonlinearity{\cdot}$ as follows
\begin{align}\label{eq:nonlinear}
\outputlaytensorentry\shiftweightfeatnot{\shiftind}{\nodeind}{\featind}\layernot{\layerind}\define\nonlinearity
{\linoutputlaytensorentry\shiftweightfeatnot{\shiftind}{\nodeind}{\featind}\layernot{\layerind}}.
\end{align}
Collecting all the elements in \eqref{eq:nonlinear},
we obtain the output of the $\layerind$-th layer 
$\outputlaytensor\layernot{\layerind}$. A common choice for 
$\nonlinearity{\cdot}$ is the rectified linear unit (ReLU), i.e. 
$\nonlinearity{c}=\text{max}(0,c)$ \cite{goodfellow2016deep}.

Hence, the main task is to define a linear transformation that maps 
$\outputlaytensor\layernot{\layerind-1}$ to $ 
\linoutputlaytensor\layernot{\layerind}$ and is tailored to our problem setup. 
Traditional convolutional NNs (CNNs) typically consider a small number of 
trainable weights and 
then generate the linear 
output as a convolution of the input with these 
weights~\cite{goodfellow2016deep}. The convolution combines values of close-by 
inputs (consecutive time instants, or neighboring pixels) and thus extracts 
information of local neighborhoods. GNNs have generalized CNNs to operate on 
graph data by replacing the convolution with a graph filter whose parameters 
are also learned~\cite{bronstein2017geometric}. This preserves locality, 
reduces the degrees of freedom of the transformation, and leverages the 
structure of the graph.

To that end, we first consider 
a step that combines linearly the information within a {graph} 
neighborhood. Since the neighborhood depends on the particular relation 
\eqref{eq:neighborhood}, we  obtain for the $i$-th relation 
\begin{align}
\diffusedfeatvec\nodeshiftnot{\nodeind}{\shiftind}\layernot{\layerind}
\define
\sum_{\nodeindp\in\neighborhoodset_\nodeind^{(\shiftind)}} \shiftentry\nodenodeshiftnot{\nodeind}{\nodeindp}{\shiftind}
\outputlayvec\nodeshiftnot{\nodeindp}{\shiftind}\layernot{\layerind-1}.
\label{eq:sem}
\end{align}
While the entries of 
$\diffusedfeatvec\nodeshiftnot{\nodeind}{\shiftind}\layernot{\layerind}$ depend 
only on the one-hop neighbors of $n$ (one-hop diffusion), the successive 
application of this operation across layers will increase the reach of the 
diffusion, spreading the information across the network. An alternative to 
account for multiple hops is to apply successively \eqref{eq:sem} 
within one layer, which is left as future work. Next, to learn features in the graph data, we combine the entries in 
$\diffusedfeatvec\nodeshiftnot{\nodeind}{\shiftind}\layernot{\layerind}$ using (trainable) parameters as follows 
\begin{align}\label{eq:linconv}\linoutputlaytensorentry\shiftweightfeatnot{\shiftind}{\nodeind}{\featind}\layernot{\layerind}\define&\sum_{\shiftindp=1}^\nbrshifts\weightshiftentry\shiftshiftfeatnot{\shiftind}{\shiftindp}{\featind}\layernot{\layerind}{\diffusedfeatvec\nodeshiftnot{\nodeind}{\shiftind}\layernot{\layerind}}\transpose\weightcol\layernot{\layerind}\nodeshiftfeatnot{\nodeind}{\shiftindp}{\featind},
\end{align}
where  the  $\nbrlayfeatures\layernot{\layerind-1}\times 1$ vector 
$\weightcol\nodeshiftfeatnot{\nodeind}{\shiftindp}{\featind} 
\layernot{\layerind}$ mixes the features and 
$\weightshiftentry\shiftshiftfeatnot{\shiftind} 
{\shiftindp}{\featind}\layernot{\layerind}$ mixes the outputs at different 
graphs. The 
$\nbrlayfeatures\layernot{\layerind-1}\times\nbrnodes\times\nbrshifts\times\nbrlayfeatures\layernot{\layerind}$
 tensor $\weighttensor\layernot{\layerind}$ collects the feature mixing weights 
$\{\weightcol\nodeshiftfeatnot{\nodeind}{\shiftindp}{\featind}\layernot{\layerind}\}$,
 while the 
$\nbrshifts\times\nbrshifts\times\nbrlayfeatures\layernot{\layerind}$ tensor 
$\weightshifttensor\layernot{\layerind}$ collects the graph mixing weights 
$\{\weightshiftentry\shiftshiftfeatnot{\shiftind} 
{\shiftindp}{\featind}\layernot{\layerind}\}$. Another key contribution of this 
paper is the consideration of $\weightshifttensor$ as a training parameter, 
which endows the GNN with the ability of learning how to mix (combine) the 
different relations encoded in the multi-relational graph. Clearly, if 
prior 
information on the dependence among relations exists, this can be used to 
constrain the structure $\weightshifttensor$ (e.g., by imposing to be diagonal 
or sparse). Upon collecting all the scalars 
$\{\linoutputlaytensorentry\shiftweightfeatnot{\shiftind}{\nodeind}{\featind}\layernot{\layerind}\}$
 in  the $\nbrshifts\times\nbrnodes\times\nbrlayfeatures\layernot{\layerind}$ 
tensor
$\linoutputlaytensor\layernot{\layerind}$, we summarize  \eqref{eq:sem}, \eqref{eq:linconv} as follows 
\begin{align}
\label{eq:lingnn}
\linoutputlaytensor\layernot{\layerind}&\define
\layfunc(\outputlaytensor\layernot{\layerind-1};
\paramhidvec\layernot{\layerind}),\;\;\text{where}\\
\label{eq:param}
\paramhidvec\layernot{\layerind}&\define[\text{vec}(\weighttensor\layernot{\layerind});\text{vec}(\weightshifttensor\layernot{\layerind})]\transpose.
\end{align}
\noindent \textbf{Recurrent GNN layer:}
Successive application of $L$ GNN layers diffuses the input $\datamatrix$ 
across the $L$-hop graph neighborhood, cf.~\eqref{eq:sem}. However, the exact 
size of the relevant neighborhood is not always known a priori.  
To endow our architecture with increased flexibility, we 
propose a recurrent GNN (GRNN) layer that inputs 
$\datamatrix$ at each $\layerind$ and, thus, captures multiple types of 
diffusion. 
Hence, the linear operation in \eqref{eq:lingnn} is replaced by the recurrent (autoregressive) 
linear tensor mapping \cite[Ch. 10]{goodfellow2016deep}
\begin{align}
\linoutputlaytensor\layernot{\layerind}\define
\layfunc(\outputlaytensor\layernot{\layerind-1};
\paramhidvec\layernot{\layerind})+
\layfunc(\datatensor;
\paraminpvec\layernot{\layerind})
\label{eq:recurrentlayer}
\end{align}
where $\paraminpvec\layernot{\layerind}$ encodes trainable parameters, cf. 
\eqref{eq:param}.  When viewed as a transformation from  $\datatensor$ to
$\linoutputlaytensor\layernot{\layerind}$, the operator in  \eqref{eq:recurrentlayer} 
implements a broader class of graph diffusions than the one in \eqref{eq:lingnn}. If 
$l=3$ for example, then the first summand in \eqref{eq:recurrentlayer} is a 1-hop diffusion of a signal that corresponded to a $2$-hop (nonlinear) diffused version of 
$\datamatrix$ while the second summand diffuses $\datamatrix$ in one hop. At a more intuitive level, the presence of the second summand also guarantees that the impact of $\datamatrix$ in the output does not vanishes as the number of layers grow. 
The autoregressive mapping in \eqref{eq:recurrentlayer} allows the architecture 
to further model time-varying inputs and labels, which motivates our future 
work towards predicting dynamic processes over multi-relational graphs\footnote{The recursive feed of  $\datamatrix$ is also known as a skip connection~\cite{mao2016image}.}. 
\subsection{Initial and final layers}
The operation of the first and last layers is very simple. Regarding layer $l=1$, the input $\outputlaytensor\layernot{0}$ is defined using $\datatensor$ as
\begin{align}\label{eq:input_first_layer}
\outputlayvec\nodeshiftnot{\nodeind}{\shiftind} \layernot{0}=\featvec_\nodeind \;\;\text{for}\;\; \text{all}\;\; (n,i).
\end{align}
On the other hand, the output of our graph architecture is obtained by taking the output of the layer $l=L$ and applying 
\begin{align}\label{eq:output}
        \predictionmat\define\outlayfun(\outputlaytensor
        \layernot{\nbrlayers};\paramoutvec),
        \end{align}
where $\outlayfun(\cdot)$ is a nonlinear function, $\predictionmat$ is an $\nbrnodes\times\nbrclasses$ matrix, 
$\predictionmatentry_{\nodeind,k}$ represents the probability that $\labelentry_\nodeind=k$, and $\paramoutvec$ are trainable parameters. The function $\outlayfun(\cdot)$ depends on the specific application, with the normalized exponential function (softmax) being a popular choice for classification problems. 

\begin{figure}
    \centering
    \includegraphics[width=\columnwidth,height=2.5cm]{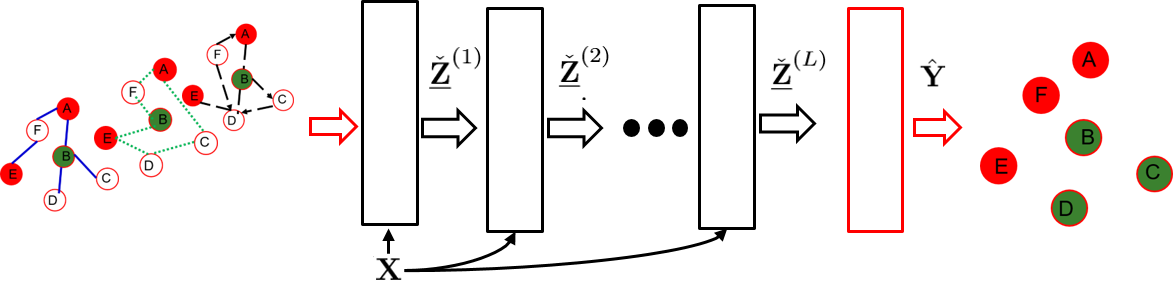}
    \caption{GRNN; $\nbrlayers$ hidden (black) and one output (red) layers}
    \label{fig:grnn}
\end{figure}

For notational convenience, the global mapping from $\datamatrix$ to 
$\predictionmat$ dictated by our GRNN architecture --i.e., by the sequential 
application of \eqref{eq:recurrentlayer}-\eqref{eq:output}-- is denoted as 
\begin{align}
 \predictionmat\define\mappingfun(\datamatrix;\{\paramhidvec\layernot{\layerind}\},\{\paraminpvec\layernot{\layerind}\},\paramoutvec),
\end{align} 
and represented in the block diagram depicted in Fig.~\ref{fig:grnn}.

\subsection{Training and graph-smooth regularizers}
The proposed architecture depends on the weights in 
\eqref{eq:recurrentlayer} and \eqref{eq:output}. We estimate these weights by 
minimizing 
the discrepancy between the estimated labels and the given ones. Hence, we 
arrive at the following minimization objective
\begin{align}\label{eq:trainobj}    \min_{\{\paramhidvec\layernot{\layerind}\},\{\paraminpvec\layernot{\layerind}\},\paramoutvec}&
\mathcal{L}_{tr} (\predictionmat,\labelmat) +\regpargraphsmooth\sum_{\shiftind=1}^\nbrshifts\text{Tr}(\predictionmat\transpose\shiftmat_\shiftind\predictionmat)
\nonumber\\
+&\regparsmooth\regfun(\{\paramhidvec\layernot{\layerind}\},\{\paraminpvec\layernot{\layerind}\})
+\regparsparse \sum_{\layerind=1}^\nbrlayers\|\weightshifttensor\layernot{\layerind}\|_1\nonumber\\
\text{s.t.}&~~
\predictionmat=\mappingfun(
\datamatrix;\{\paramhidvec\layernot{\layerind}\},\{\paraminpvec\layernot{\layerind}\},\paramoutvec). 
\end{align}
In our classification setup, a sensitive choice for the fitting cost is to use  $\mathcal{L}_{tr} (\predictionmat,\labelmat)\define-\sum_{\nodeind\in\labeledset}\sum_{k=1}^\nbrclasses 
\labelmatentry_{\nodeind k}\ln{\predictionmatentry_{\nodeind k}}$ the cross-entropy loss function over the labeled examples.

Note also that three regularizers have been considered. The first (graph-based) 
regularizer promotes smooth label estimates over the graphs 
\cite{ioannidis2018kernellearn}, and  $\regfun(\cdot)$ is an $\mathcal{L}_2$ 
norm over the GRNN parameters that is typically used to avoid overfitting 
\cite{goodfellow2016deep}. Finally, the $\mathcal{L}_1$ norm in the third 
regularizer promotes learning sparse mixing coefficients and, hence, promotes 
activating only a subset of relations at each $\layerind$.  The backpropagation 
algorithm~\cite{rumelhart1986learning}  is employed to minimize 
\eqref{eq:trainobj}. The computational complexity of evaluating 
\eqref{eq:recurrentlayer} scales linearly with the number of nonzero entries in 
$\shifttensor$ (edges), cf. \eqref{eq:sem}.

To recap, while most of the works in the GNN literature use a single graph with 
one type of diffusion \cite{bronstein2017geometric,kipf2016semi}, we have 
proposed a (recurrent) multi-relational GNN architecture that: adapts to each 
graph with $\weightshifttensor$; uses a simple but versatile recurrent  
tensor  mapping \eqref{eq:recurrentlayer}; and 
includes several types of graph-based regularizers.

\section{Numerical tests}
We test the proposed GNN with $\nbrlayers=2$, $\nbrlayfeatures\layernot{1}=64$, 
and  $\nbrlayfeatures\layernot{2}=\nbrclasses$. The regularization parameters 
$\{\regpargraphsmooth,\regparsmooth,\regparsparse\}$ are chosen based on the 
performance of the GRNN in the validation set for each experiment. For the 
training stage, an ADAM optimizer with learning rate 0.005 was employed 
\cite{kingma2015adam}, for 300 epochs\footnote{An epoch is a cycle through all 
the training examples} with early stopping at 60 epochs\footnote{Training stops 
if the validation loss does not decrease for 60 epochs}. The multiple layers of 
the graph in this case are formed using $\kappa$-nearest neighbors graphs for 
different values of $\kappa$ (i.e., different number of neighbors). This method 
computes the link between $\nodeind$ and $\nodeindp$ based on the Euclidean 
distance of their features $\|\featvec_\nodeind-\featvec_\nodeindp\|_2^2$.  The 
simulations were run using TensorFlow~\cite{abadi2016tensorflow} and the code 
is available 
online\footnote{https://sites.google.com/site/vasioannidispw/github}.
\begin{figure*}[t]
    \centering
    \begin{subfigure}[b]{\columnwidth}
    \centering{
%
%
\begin{tikzpicture}

\begin{axis}[%
width=0.956\mywidth,
height=0.987\myheight,
at={(0\mywidth,0\myheight)},
scale only axis,
xlabel style={font=\color{white!15!black}},
xlabel={SNR of $\featperturbation$},
xmin=0.2,
xmax=25,
xmode=log,xtick={0.2,1,5,25,125},xticklabels={0.2,1,5,25,125},
ylabel style={font=\color{white!15!black}},
ylabel={Classification accuracy},
legend columns=4,
xmajorgrids,
ymajorgrids,
  cycle list name=colorlist,
grid style={dotted},ticklabel style={font=\ticklabelfontsize},
legend style={
	at={(0,1.015)}, 
	anchor=south west, legend cell align=left, align=left,
	draw=none
	, font=\legendfontsize}
]
\addplot
  table[row sep=crcr]{%
0.2	0.64\\
1	0.68\\
5 0.92\\
25	0.98\\
};
\addlegendentry{$\kappa=$2}
\addplot
  table[row sep=crcr]{%
0.2	0.79\\
1	0.80\\
5 0.96\\
25	0.98\\
};
\addlegendentry{$\kappa=$5}
\addplot
  table[row sep=crcr]{%
0.2	0.89\\
1	0.91\\
5 0.97\\
25	0.99\\
};
\addlegendentry{$\kappa=$10}
\addplot
  table[row sep=crcr]{%
0.2	0.90\\
1	0.93\\
5 0.98\\
25	0.993\\
};
\addlegendentry{$\kappa=$5,10}
\end{axis}
\end{tikzpicture}%
}
	\caption{Classification accuracy with noisy features.}
    \end{subfigure}%
    ~ 
    \begin{subfigure}[b]{\columnwidth}
    \centering{
%
%
\begin{tikzpicture}

\begin{axis}[%
width=0.956\mywidth,
height=0.987\myheight,
at={(0\mywidth,0\myheight)},
scale only axis,
xlabel style={font=\color{white!15!black}},
xlabel={SNR of $\shiftperturbation$},
xmin=0.2,
xmax=25,
xmode=log,xtick={0.2,1,5,25,125},xticklabels={0.2,1,5,25,125},
ylabel style={font=\color{white!15!black}},
ylabel={Classification accuracy},
legend columns=4,
xmajorgrids,
ymajorgrids,
  cycle list name=colorlist,
grid style={dotted},ticklabel style={font=\ticklabelfontsize},
legend style={
	at={(0,1.015)}, 
	anchor=south west, legend cell align=left, align=left,
	draw=none
	, font=\legendfontsize}
]
\addplot
  table[row sep=crcr]{%
0.2	0.50\\
1	0.52\\
5 0.53\\
25	0.67\\
125	0.98\\
};
\addlegendentry{$\kappa=$2}
\addplot
  table[row sep=crcr]{%
0.2	0.51\\
1	0.52\\
5 0.52\\
25	0.71\\
125	0.99\\
};
\addlegendentry{$\kappa=$5}
\addplot
  table[row sep=crcr]{%
0.2	0.50\\
1	0.52\\
5 0.50\\
25	0.76\\
125	0.99\\
};
\addlegendentry{$\kappa=$10}
\addplot
  table[row sep=crcr]{%
0.2	0.51\\
1	0.56\\
5 0.60\\
25	0.79\\
125	0.999\\
};
\addlegendentry{$\kappa=$5,10}
\end{axis}
\end{tikzpicture}%
}
	\caption{Classification accuracy with noisy graphs.}
    \end{subfigure}\\%
    \centering
    \begin{subfigure}[b]{\columnwidth}
    \centering{
%
%
\begin{tikzpicture}

\begin{axis}[%
width=0.956\mywidth,
height=0.987\myheight,
at={(0\mywidth,0\myheight)},
scale only axis,
xlabel style={font=\color{white!15!black}},
xlabel={SNR of $\featperturbation$},
xmin=0.2,
xmax=25,
xmode=log,xtick={0.2,1,5,25,125},xticklabels={0.2,1,5,25,125},
ylabel style={font=\color{white!15!black}},
ylabel={Classification accuracy},
legend columns=4,
xmajorgrids,
ymajorgrids,
  cycle list name=colorlist,
grid style={dotted},ticklabel style={font=\ticklabelfontsize},
legend style={
	at={(0,1.015)}, 
	anchor=south west, legend cell align=left, align=left,
	draw=none
	, font=\legendfontsize}
]
\addplot
  table[row sep=crcr]{%
0.2	0.6\\
1	0.64\\
5 0.75\\
25	0.84\\
};
\addplot
  table[row sep=crcr]{%
0.2	0.74\\
1	0.76\\
5 0.82\\
25	0.83\\
};

\addplot
  table[row sep=crcr]{%
0.2	0.76\\
1	0.76\\
5 0.78\\
25	0.81\\
};
\addplot
  table[row sep=crcr]{%
0.2	0.78\\
1	0.80\\
5 0.83\\
25	0.85\\
};
\end{axis}
\end{tikzpicture}%
}
	\caption{Classification accuracy with noisy features.}
    \end{subfigure}%
    ~ 
    \begin{subfigure}[b]{\columnwidth}
    \centering{
%
%
\begin{tikzpicture}

\begin{axis}[%
width=0.956\mywidth,
height=0.987\myheight,
at={(0\mywidth,0\myheight)},
scale only axis,
xlabel style={font=\color{white!15!black}},
xlabel={SNR of $\shiftperturbation$},
xmin=0.199,
xmax=25.001,
xmode=log,
xtick={0.2,1,5,25,125},
xticklabels={0.2,1,5,25,125},
ylabel style={font=\color{white!15!black}},
ylabel={Classification accuracy},
legend columns=4,
xmajorgrids,
ymajorgrids,
  cycle list name=colorlist,
grid style={dotted},ticklabel style={font=\ticklabelfontsize},
legend style={
	at={(0,1.015)}, 
	anchor=south west, legend cell align=left, align=left,
	draw=none
	, font=\legendfontsize}
]
\addplot
  table[row sep=crcr]{%
0.2	0.58\\
1	0.61\\
5 0.65\\
25	0.75\\
125	0.84\\
};
\addplot
  table[row sep=crcr]{%
0.2	0.59\\
1	0.58\\
5 0.69\\
25	0.72\\
125	0.84\\
};
\addplot
  table[row sep=crcr]{%
0.2	0.60\\
1	0.62\\
5 0.63\\
25	0.74\\
125	0.85\\
};
\addplot
  table[row sep=crcr]{%
0.2	0.58\\
1	0.63\\
5 0.64\\
25	0.73\\
125	0.83\\
};
\end{axis}
\end{tikzpicture}%
}
	\caption{Classification accuracy with noisy graphs.}
    \end{subfigure}%
	\caption{Classification accuracy on synthetic (a), (b) and ionosphere (c), (d). 
	}
\label{fig:robust}
\end{figure*}
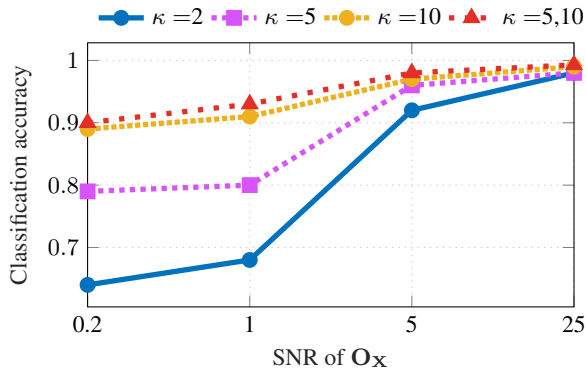
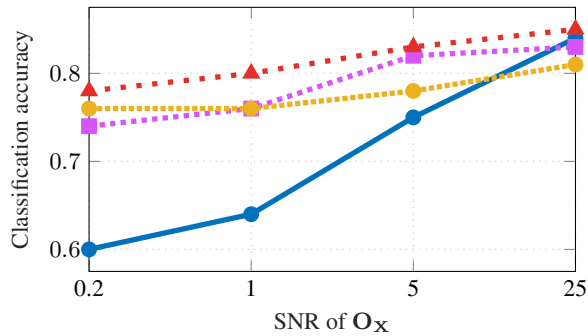
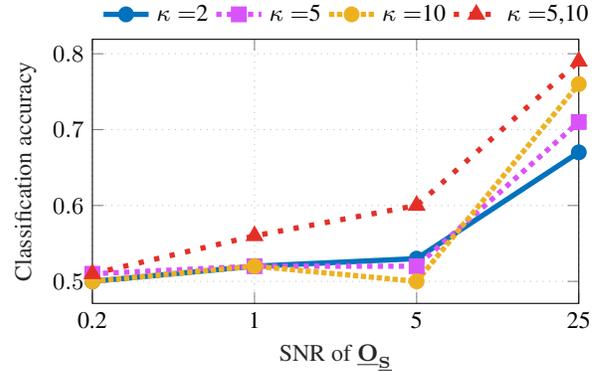
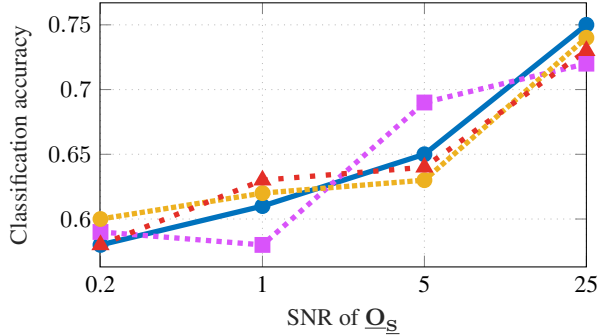
\subsection{Robustness of GRNN}
This section reports the performance of the proposed architecture under perturbations. Oftentimes, the available topology and feature vectors might be corrupted  (e.g. due to privacy concerns or because adversarial social users manipulate the data to sway public opinion). In those cases, the observed $\shifttensor$ and $\datamatrix$ can be modeled as 
\begin{align}
    \label{eq:toppertub}
    \shifttensor=&\shifttensor_{tr}+\shiftperturbation\\
     \datamatrix=&\datamatrix_{tr}+\featperturbation\label{eq:featpertub}.
\end{align}
where $\shifttensor_{tr}$ and $\datamatrix_{tr}$ represent the \textit{true} topology and features and $\shiftperturbation$ and $\featperturbation$ denote the corresponding additive perturbations. We draw  $\shiftperturbation$ and $\featperturbation$  from a zero mean white Gaussian distribution with specified signal to noise ratio (SNR). The robustness of our method is tested in two datasets: i)  
A synthetic dataset of $\nbrnodes=1000$ points that belong to $\nbrclasses=2$ 
classes generated as
$\featvec_{\nodeind}\in\rfield^{\nbrinpfeatures\times1}\sim\mathcal{N}(\mu,0.4)$,
$\nodeind=1,\ldots,1000$ with $\nbrinpfeatures=10$ and $\mu=0,1$ corresponding 
to the different classes.
ii) The ionosphere dataset, which contains $\nbrnodes=351$ data points with $\nbrinpfeatures=34$ features that belong to $\nbrclasses=2$ classes \cite{Dua:2017}. We generate $\kappa$-nearest neighbors graphs by varying $\kappa$, and observe  $|\labeledset|=200$ and $|\labeledset|=50$ nodes uniformly at random.

With this simulation setup, we test the different GRNNs in SSL for increasing 
the SNR of $\shiftperturbation$ (Figs. \ref{fig:robust}a, \ref{fig:robust}c) 
and $\featperturbation$ (Figs. \ref{fig:robust}b, \ref{fig:robust}d). We deduce 
from the classification performance of our method in Fig. \ref{fig:robust} that 
multiple graphs leads to learning more robust representations of the data, 
which testifies to the merits of proposed multi-relational architecture.

\subsection{Classification of citations graphs}
We test our architecture with three citation network datasets~\cite{sen2008collective}. The citation graph is denoted as $\shiftmat_0$, its nodes correspond to different documents from the same scientific category, and $\shiftentry\nodenodeshiftnot{\nodeind}{\nodeindp}{0}=1$ implies that paper $\nodeind$ cites paper $\nodeindp$. Each document $\nodeind$ is associated with a feature vector $\featvec_\nodeind$ that measures the frequency of a set of words, as well as with a label $\labelentry_\nodeind$ that indicates the document's subcategory. ``Cora'' contains papers related to machine learning, ``Citeseer'' includes papers related to computer and information science, while ``Pubmed''  contains biomedical papers, see also Table \ref{tab:citation}.
\begin{table}[]
\hspace{0cm}
    \centering
    \begin{tabular}{c c c c c}
    \textbf {Dataset} & \textbf {Nodes} $\nbrnodes$ & \textbf {Classes}
    $\nbrclasses$ & \textbf {Features} $\nbrinpfeatures$ &  $|\labeledset|$\\
    \hline
       Cora  & 2,708 & 7 & 1,433 & 140\\
       Citeseer & 3,327 & 6 & 3,703 & 120  \\
       Pubmed & 19,717 & 3 & 500 & 30
    \end{tabular}
    \caption{Citation datasets}
    \label{tab:citation}
\end{table}

To facilitate comparison, we reproduce the same experimental setup than in 
\cite{kipf2016semi}, i.e., the same split of the data in train, validation, and 
test sets. We test two architectures: a) a GRNN using only the original 
citation graph $\shiftmat_0$ (and, hence, with $\nbrshifts=1$); and b) a 
multirelational GRNN that uses an extra 1-nearest neighbor graph (so that 
$\nbrshifts=2$).    Table~\ref{tab:result} reports the classification accuracy 
of various SSL methods. It is observed that: i) GNN approaches (ours, as well 
as \cite{kipf2016semi}) outperform competing alternatives, ii) our GRNN schemes 
always outperform \cite{kipf2016semi} (either with $I=1$ or $I=2$). This 
illustrate the potential benefits of the recurrent feed in 
\eqref{eq:recurrentlayer} --not used in \cite{kipf2016semi}-- as well as the 
use of multi-relational graphs. 
\begin{table}[]
    \centering
    \begin{tabular}{c c c c}
    \textbf{Method} & \textbf{Citeseer} & \textbf{Cora} & \textbf{Pubmed}\\
    \hline
    ManiReg~\cite{belkin2006manifold} & 60.1& 59.5 & 70.7 \\
    \hline
    SemiEmb~\cite{weston2012deep} & 59.6 & 59.0 & 71.7 \\
        \hline
    LP~\cite{zhu2003semi} & 45.3 & 68.0 & 63.0 \\
        \hline
    Planetoid~\cite{yang2016revisiting} & 64.7 & 75.7 & 77.2 \\
        \hline
    GCN~\cite{kipf2016semi} & 70.3 & 81.5 & 79.0 \\
        \hline
    GRNN\footnote{$\regpargraphsmooth=5\times10^{-5},\regparsmooth=5\times10^{-5},\regparsparse=10^{-4}$, dropout rate=0.9}  & \textbf{70.8}
    &  \textbf{82.8}     
    & \textbf{79.5}\\
        \hline
    GRNN (multi-relational)\footnote{$\regpargraphsmooth=2\times10^{-6},\regparsmooth=5\times10^{-5},\regparsparse=10^{-4}$, dropout rate=0.9} & \textbf{70.9}
    &\textbf{81.7} 
    & \textbf{79.2}
    \end{tabular}
    \caption{Classification accuracy for citation datasets}
    \label{tab:result}
\end{table}

\section{Conclusions}
This paper put forth a novel deep learning framework for SSL that utilized an 
autoregressive multi-relational graphs architecture to sequentially process the 
input data.  Instead of committing a fortiori to
a specific type of diffusion, our novel GRNN learns the diffusion pattern that 
best fits the data. The proposed architecture is able to handle scenarios where nodes 
engage in multiple relations, can be used to reveal the structure of the data, 
and is computationally affordable, since the number of operations scaled 
linearly with respect to the number of graph edges.
Our approach achieves state-of-the-art classification
results on graphs when nodes are accompanied 
by feature vectors.  Future research includes investigating robustness to 
adversarial topology perturbations, predicting time-varying labels,  and 
designing of pooling operators. 
\newpage

\bibliographystyle{IEEEtran}
\bibliography{my_bibliography}
\noindent
\end{document}